\documentclass[10pt,twocolumn,letterpaper]{article}

\usepackage[pagenumbers]{cvpr} 
\usepackage[table,xcdraw]{xcolor}
\usepackage{graphicx}
\usepackage{amsmath}
\usepackage{amssymb}
\usepackage{booktabs}

\usepackage[pagebackref,breaklinks,colorlinks]{hyperref}

\usepackage[capitalize]{cleveref}
\crefname{section}{Sec.}{Secs.}
\Crefname{section}{Section}{Sections}
\Crefname{table}{Table}{Tables}
\crefname{table}{Tab.}{Tabs.}

\def\cvprPaperID{8561} 
\def\confName{CVPR}
\def\confYear{2022}

\begin{document}

\title{ExpNet: A unified network for Expert-Level Classification}

\author{Junde Wu\\
Healthcare Group, Baidu\\
{\tt\small jundewu@ieee.org}
\and
Huihui Fang\\
Healthcare Group, Baidu
\and
Yu Zhang\\
Harbin Institute of Technology
\and
Yehui Yang\\
Healthcare Group, Baidu
\and
Haoyi Xiong\\
Big Data Lab, Baidu
\and
Huazhu Fu\\
IHPC, A*STAR
\and
Yanwu Xu\\
Healthcare Group, Baidu\\
{\tt\small ywxu@ieee.org}
}
\maketitle

\begin{abstract}
Different from the general visual classification, some classification tasks are more challenging as they need the professional categories of the images. In the paper, we call them expert-level classification. Previous fine-grained vision classification (FGVC) has made many efforts on some of its specific sub-tasks. However, they are difficult to expand to the general cases which rely on the comprehensive analysis of part-global correlation and the hierarchical features interaction. In this paper, we propose Expert Network (ExpNet) to address the unique challenges of expert-level classification through a unified network. In ExpNet, we hierarchically decouple the part and context features and individually process them using a novel attentive mechanism, called Gaze-Shift. In each stage, Gaze-Shift produces a focal-part feature for the subsequent abstraction and memorizes a context-related embedding. Then we fuse the final focal embedding with all memorized context-related embedding to make the prediction. Such an architecture realizes the dual-track processing of partial and global information and hierarchical feature interactions. We conduct the experiments over three representative expert-level classification tasks: FGVC, disease classification, and artwork attributes classification. In these experiments,  superior performance of our ExpNet is observed comparing to the state-of-the-arts in a wide range of fields, indicating the effectiveness and generalization of our ExpNet. The code will be made publicly available.



\end{abstract}
\vspace{-3pt}
\section{Introduction}
\label{sec:intro}

\begin{figure}[!t]
\centering
\includegraphics[width=\linewidth]{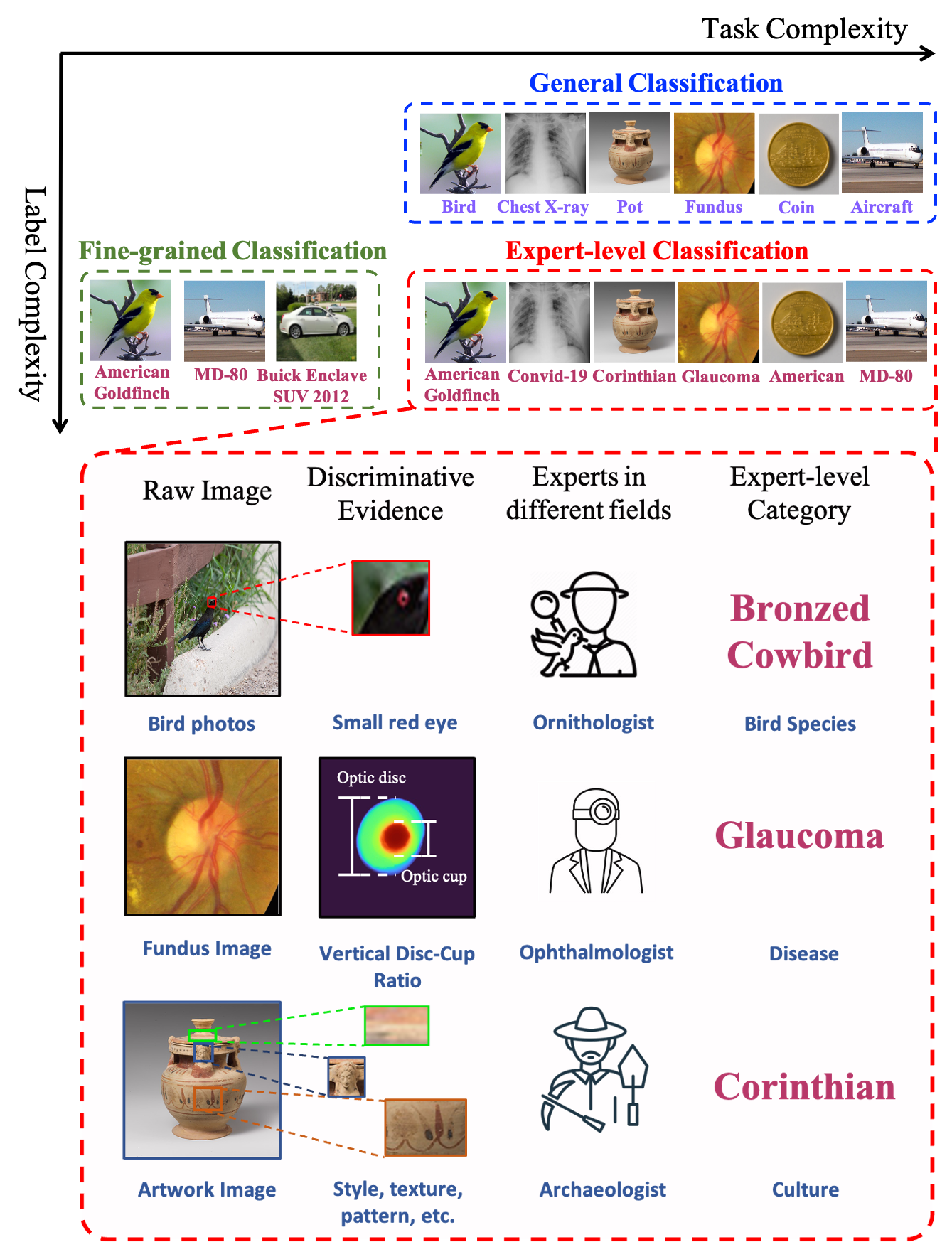}
\caption{Expert-level classification is more challenging than general and fine-grained vision classification as it faces both Label Complexity (professional labels) and Task Complexity (various tasks). We show the unique challenges of expert-level classification by three representative examples. In bird species classification, the detailed parts are important, e.g., the tiny red eye is discriminative evidence of bronzed cowbird. In glaucoma classification, the global structure relationship is important, e.g., the large vertical disc-cup ratio is a major biomarker for glaucoma. In the artwork attributes classification, the interaction of different features is important, e.g., to recognize the culture of the pot needs a comprehensive analysis of the style, texture, and material of the object.}
\label{fig:facial}
\vspace{-15pt}
\end{figure}

Deep neural network (DNN) has made great progress in computer vision tasks. As a fundamental task in this field, image classification has been studied by a large number of methods. Thanks to the effort of these researches, image classification has achieved a very high level, which is even comparable to humankind on some of the tasks~\cite{geirhos2017comparing}. 

In human society, some visual classification tasks are even challenging for ordinary people. These tasks can only be completed by a small part of our humankind who have experienced a long-term professional trainee, whom we call experts. For example, only ornithologists can recognize the bird species from a wild bird photo, only ophthalmologists can accurately diagnose glaucoma from the fundus images, and only archaeologists can tell the culture of an artwork from a picture of it. Since the expert-level categories are semantically more complex than the general ones, these tasks will also be more challenging for deep learning models. 
Previous Fine-grained Vision Classification (FGVC)~\cite{zhao2017survey} delves into some of its specific tasks, e.g., to recognize the bird species, car brands, or aircraft types. These tasks share the same characteristic that some parts or details are discriminative for the classification, which FGVC methods can often take advantage of. But not all the tasks with expert-level categories have the same characteristic, and can be well addressed by FGVC methods. Just as the taxonomy shown in Fig.~\ref{fig:facial}, compared with general classification, FGVC predicts more complex labels but can only process limited types of tasks.

This paper aims to address a kind of classification problem with a larger scope of FGVC, which we call expert-level classification. We define expert-level classification as a classification task in which the target category needs to be further inferred or analyzed from the salient objects in the images. A wide range of tasks fall into this definition, like recognizing the bird species from the bird, detecting the diseases from the lesions/tissues, discovering the culture of the artworks, and recognizing the artists of the paintings. Facing both task complexity and label complexity, expert-level classification is more challenging than general classification tasks and FGVC.

To be specific, the challenges of expert-level classification can be concluded as three points. First, the devils are in the details. Like the bird case shown in Fig.~\ref{fig:facial}, a tiny detail of the image may decide the category of the whole image. Second, not only the local details, but the global structures also matter. For example, as the fundus case in Fig.~\ref{fig:facial}, a large vertical optic cup and disc ratio (vCDR) of the fundus image is a major biomarker indicating glaucoma. Finally, feature-level interactions are also important. As the artwork case in Fig.~\ref{fig:facial}, 
a comprehensive analysis of various features like painting style, object textures, and patterns is the key to recognize the Corinthian culture of the pot. 

In order to address the unique challenges in expert-level classification, we propose a novel network to decouple and individually process the focal and global information in a hierarchical manner. The basic idea is to progressively zoom in the attractive parts and memorize the context features in each stage. In the end, we fuse the final focal feature and all the memorized context-related features to make the final decisions. 
In the implementation, we design Gaze-Shift to produce a focal-part feature map and a Context Impression embedding in each stage. In Gaze-Shift, we first use the proposed Neural Firing Fields (NeFiRF) to split focus parts and context parts of the given feature map.
Then we use convolution layers to extract the focal parts, and global Cross Attention (CA)\cite{chen2021crossvit} to model the context correlation. The focal-part feature will be sent to the next stage, and Context Impression will be stripped out for the final fusion. Such an architecture models part-global trade-offs and hierarchical feature interactions to overcome the unique challenges of expert-level classification. We verify the effectiveness of ExpNet on three representative tasks of expert-level classification, which are FGVC, disease classification, and artwork attributes classification. The experiments show ExpNet outperforms state-of-the-art (SOTA) on several mainstream benchmarks with solid generalization ability. Moreover, the intermediate results produced by ExpNet show a close relationship with salient object shape and location, which yields competitive performance in weakly-supervised segmentation/localization.


The contributions of the paper can be concluded as follow. 1). We verify that a significant and challenging vision classification task, i.e., expert-level classification, can be solved by a unified deep learning framework. 2). We propose ExpNet with Gaze-Shift to individually process the focal and context features by different parameters and architectures in a hierarchical manner, which helps to overcome the unique challenges of expert-level classification. 3). We propose NeFiRF in Gaze-Shift to group the features in the frequency space with spatial constraints. 4). We achieve SOTA on three representative expert-level classification tasks compared with both task-specific and general classification methods. 5). We achieve competitive performance on weakly-supervised segmentation and localization.






\begin{figure*}[!t]
\centering
\includegraphics[width=0.85\linewidth]{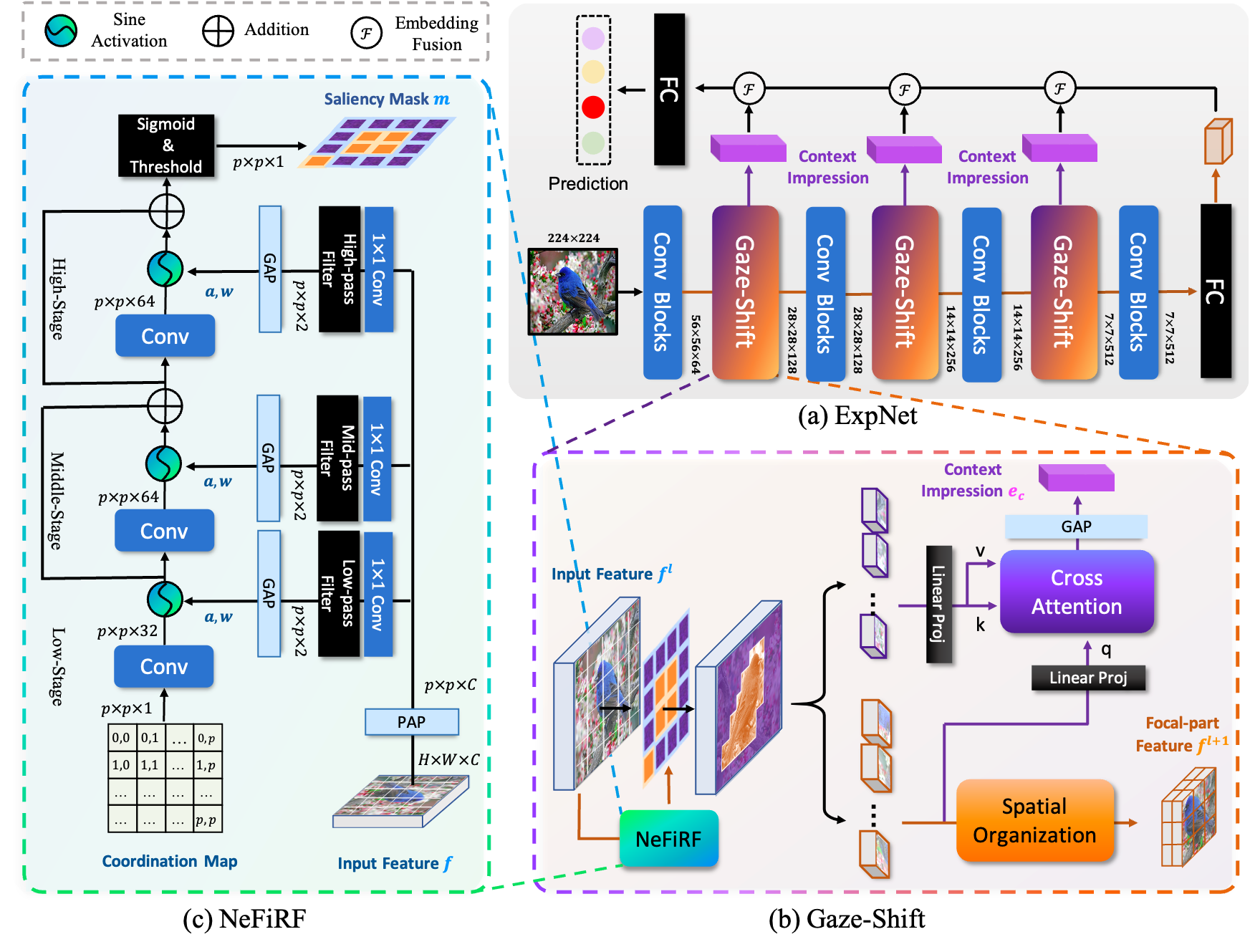}
\caption{
An illustration of our ExpNet framework, which starts from (a) an overview pipeline using ResNet34 backbone as the example, and continues with zoomed-in diagrams of individual modules, including (b) Gaze-Shift, and (c) Neural Firing Fields (NeFiRF).}
\vspace{-15pt}
\label{fig:framework}
\end{figure*}

\vspace{-3pt}
\section{Related Work}
Fine-grained visual categorization (FGVC) aims to distinguish subordinate categories within entry-level categories, which can be seen as a special case of expert-level classification. Previous FGVC strategies include exploring multi-level feature interaction~\cite{luo2019cross}, locating discriminative parts~\cite{zhang2019learning,ge2019weakly} and identifying the subtle difference through pair-wise learning~\cite{dubey2018pairwise,ji2020attention} or patch-level shuffling~\cite{chen2019destruction,du2020fine}. Among them, the part-based methods are most similar to our idea. The previous CNN-based implementations\cite{zhang2019learning, ge2019weakly} first predict the part localizations and then classify based on the selected regions. Recent vision transformer (ViT)~\cite{dosovitskiy2020image} based implementations~\cite{he2022transfg, zhu2022dual}
adopt self-attention mechanism to model the local-global interaction. But they often select the parts through the response of the features, in which the selection and classification are highly coupled. In addition, they equally process the global and part information at each level, which dilutes the importance of the discriminative features. 

Besides FGVC, the regional attention mechanism itself is a popular technique that has been widely studied. A basic idea is to first localize the attractive parts, and then classify based on the region of interest~\cite{bajwa2019two, fu2018disc}. The alternative strategies include enhancing the features with self-transformation~\cite{woo2018cbam,tian2020attention,chen2017sca} or using localization information~\cite{wu2022seatrans,wu2020leveraging}. However, these space-based methods are likely to cause overfitting since the learnable attention maps are strongly correlated with the adopted features. 
Recently, there are studies~\cite{rao2021global,tang2022image,lee2021fnet} show that aggregating/filtering the features in frequency domain space is more efficient and robust. However, they are easy to introduce high-frequency noises by completely ignoring the spatial constraint. 
\vspace{-3pt}
\section{Method}
\subsection{Motivation and ExpNet}\label{sec:hier}
Based on the research of human vision system~\cite{barghout1999differences,barghout2003global,lindeberg2013computational}, when we are recognizing an image, we will first shift the attention to a salient object, such as the bird in the picture, and also retains the impression of the context, such as the spring season, in the memory. When an expert needs to obtain more professional information from the image, he/she will repeat this process more times than ordinary people~\cite{barghout2003global}. For example, the expert will further observe the details of birds, such as wings and beaks, and retain the overall impression of birds, such as the blue color, in the memory. Later, more detailed features, such as the patterns of feathers on the wings, will be observed, and so on. Finally, the expert can integrate all the memorized information and the observation to make a comprehensive inference. 

Inspired by this biological process, we design ExpNet to hierarchically decouple the focal part and the context information, then individually process the two parts in each stage. An illustration is shown in Fig.~\ref{fig:framework} (a). Over a standard ResNet~\cite{he2016deep} backbone, we propose a module named Gaze-Shift between two Convolution Blocks to decompose the $l$ stage feature map $f^{l}$ to the focal-part feature $f^{l+1}$ and a context-related embedding $e_{c}^{l}$, named Context Impression. The focal-part feature will be sent to the next stage and Context Impression will be memorized. Repeating this way, the local but discriminative parts will be progressively abstracted to high-level semantic features, while the Context Impression will be stripped out at its appropriate level to reinforce/calibrate the final decision.
Using ResNet as the backbone, we stack four of such stages to obtain a final focal embedding and three Context Impression. These embeddings are then fused (Section \ref{sec:fuse}) to obtain the final decision. The whole network is trained end-to-end by image-level categories using cross-entropy loss. 

\vspace{-3pt}
\subsection{Gaze-Shift}\label{sec:hier}
Specifically, Gaze-Shift works as which shown in Fig. \ref{fig:framework} (b). Given a feature $f\in \mathbb{R}^{H \times W \times C}$, we first pachlized the feature map with a patch size $k$. Then we use NeFiRF (Section \ref{sec:nerf}) to generate a binary patch-level map $m \in \mathbb{R}^{p \times p \times 1}$, called saliency map, where $p = H / k$ considering the common case $H$ equals to $W$. Saliency map spatially splits the patches of the feature map to focal patches and context patches. In the focus branch, the selected focal patches are spatially organized (Section \ref{sec:so}) to keep the spatial correlations. In this process, they will be downsampled and abstracted to the focal-part feature $f^{l+1} \in \mathbb{R}^{\frac{H}{2} \times \frac{W}{2} \times 2C}$. In the context branch, the focus patches and context patches are interacted through Cross Attention (CA)~\cite{chen2021crossvit} to obtain $e_{c}$. The focal parts are used as the \textit{query}, the context parts are used as the \textit{key} and \textit{value} to be interacted by the attentive mechanism (see Section \ref{sec:ca} for details). In this way, we encode the discriminative embedding from the context content considering its interaction with the focus parts. 
\vspace{-3pt}
\subsection{NeFiRF}\label{sec:nerf}
Spatially grouping and weighting the features is a well-recognized strategy in classification tasks.
The common practices include learning the spatial attention map~\cite{woo2018cbam} or measuring the feature similarity~\cite{liu2022dynamic}. However, these spatial modeling methods will likely cause overfitting as the feature maps and attention maps are strongly correlated.
Recently, processing the features in the frequency space has shown to be an efficient way to promise the generalization~\cite{rao2021global,lee2021fnet,tang2022image}.
However, directly learning the filter in the frequency space~\cite{rao2021global} will lose the spatial constraint of the feature map. Note that 
our grouping is supposed to be naturally spatial correlated, i.e., focus parts or context parts are likely to gather together spatially.
To introduce the spatial constraint in the frequency space, we generate the saliency map by a conditional Neural Fields (NeRF)~\cite{xie2022neural}, which we named Neural Firing Fields (NeFiRF). NeFiRF is a NeRF conditioned by the frequency encoding of the given feature map. As the neural networks are prone to produce similar responses given similar inputs, NeRF inputted by a coordination map will naturally constrain the output to be spatially smooth~\cite{xie2022neural}. 

An illustration of NeFiRF is shown in Fig.~\ref{fig:framework} (c). Specifically, NeFiRF generates a binary patch-level map $m$ conditioned by the inputted feature map $f$. The map assigns each patch of the feature map to ‘focus’ or ‘context’. The main architecture of NeFiRF is a 2-D NeRF, which predicts the value of the position from the inputted coordination map.
Different from the original implementation of NeRF, we apply convolution layer over the input coordination map.
We build six convolution layers in total, one in the low-stage, four in the middle-stage, and another one in the high-stage. To introduce the feature frequency, we follow SIREN~\cite{sitzmann2020implicit} to insert the periodic activation function between each two convolution layers, but we control the activation through the embedding of $f$ instead of the direct learning. 
Specifically, we encode $f$ as the parameters to control the amplitude $a$ and frequency $w$ of the sine activation function $a sin(wx)$, where $x$ is the inputted feature element. In the implementation, we first apply patch-average pooling (PAP) to convert $f$ to a $p\times p\times C$ map, then encode it by $1 \time 1$ convolution to a $p \times p \times 2$ map. The two channels respectively represent the amplitude and frequency for each patch. We then control the bandwidth of each different stage to encourage coarse-to-fine generation. Following the physical concept of band-pass filters, we eliminate the top 20\% high-frequency activation to zero in the low-pass filter, the top 10\% high-frequency and low-frequency activation to zero in the middle-pass, and the top 20\% low-frequency activation to zero in the high-pass filter. After then, we use global average pooling (GAP)~\cite{lin2013network} to produce two values $a$ and $w$ to decide a unique activation. 
In NeFiRF, we adopt the skip connection over the middle and high stages to pass the low-frequency information to the higher levels. The Sigmoid function and 0.5 thresholding are adopted on the last layer to produce a binary $p \times p$ saliency map. NeFiRF will not share weights in different stages.



\vspace{-3pt}
\subsection{Details of Implementation}\label{sec:details}
\textbf{Spatial Organization}\label{sec:so}
After NeFiRF selects the focal patches, we reorganize and abstract them to the next stage feature map while maintaining their original spatial correlation. The process is implemented by padding, max pooling, and deformable convolution~\cite{dai2017deformable}. 
Specifically, we first arrange the selected focal patches according to their previous positions and adopt padding following 2-stride max pooling to downsample the feature map to $\frac{H}{2}\times \frac{H}{2} \times C$.
Then, we apply a deformable convolution on the feature map to keep the same scale but double the channels. Deformable convolution learns the offsets together with the convolution parameters. For each convolution kernel, the offset helps it abstract only the informative positions and ignore the blank ones. In this way, it actually enlarges the selected focus features to the blank positions. More details and experiments about the spatial organization are provided in supplementary materials.

\textbf{Cross Attention}\label{sec:ca}
We use Cross Attention (CA)~\cite{chen2021crossvit} to model the interaction between the focus and context features in each stage. 
Different from its original setting, we do not use the class token for the classification. Instead, only embeddings (flattened from the feature maps) are interacted through the attention mechanism. GAP is applied on the produced results to obtain the final embedding. 
In addition, we use conditional position encoding~\cite{chu2021conditional} to encode the position information. The flattened focus embedding and flattened context embedding will be concatenated as the condition for positional encoding learning. The produced encoding is split to the focus and context parts based on the positions, and respectively added to the focus embedding and context embedding.

\textbf{Embedding Fusion}\label{sec:fuse}
We tried several ways to fuse the embedding for the classification, including concatenation + multilayer perceptron (MLP), MLP + multiplication, MLP + addition, dynamic MLP~\cite{yang2022dynamic} and CA. The detailed implementation and experiments can be found in supplementary materials. In the experiments, we adopt CA in the large variant of the model and MLP + addition in the small variant of the model.


\vspace{-3pt}
\section{Experiments}

\begin{table*}[!t]
\centering
\resizebox{0.8\textwidth}{!}{
\begin{tabular}{c|c|c|cc|cc|cc|c}
\toprule
\hline
\multicolumn{1}{c|}{Method}     & Param       & Architecture               & \multicolumn{2}{c|}{FGVC}                                                                         & \multicolumn{2}{c|}{Medical}                                                                                         & \multicolumn{2}{c|}{Artworks}                                                                                             & Mean                        \\ \hline
&                        &                & \begin{tabular}[c]{@{}c@{}}CUB\\ (Acc\%)\end{tabular} & \begin{tabular}[c]{@{}c@{}}Air\\ (Acc\%)\end{tabular} & \begin{tabular}[c]{@{}c@{}}REFUGE2\\ (AUC\%)\end{tabular} & \begin{tabular}[c]{@{}c@{}}Convid-19\\ (Acc\%)\end{tabular} & \begin{tabular}[c]{@{}c@{}}iMet2020\\ (F2 Score\%)\end{tabular} & \begin{tabular}[c]{@{}c@{}}WikiArt\\ (Acc\%)\end{tabular} &                             \\ \hline
\multicolumn{1}{c|}{BCN~\cite{dubey2018pairwise}}    &25M           & ResNet-50           & \cellcolor[HTML]{EFEFEF}87.7                        & \cellcolor[HTML]{EFEFEF}90.3                         & 78.7                                                    & 91.0                                                      & 57.8                                                          & 72.6                                                     & 79.7                        \\
\multicolumn{1}{c|}{ACNet~\cite{ji2020attention}}   &48M           & ResNet-50           & \cellcolor[HTML]{EFEFEF}88.1                        & \cellcolor[HTML]{EFEFEF}92.4                         & 80.5                                                    & 93.2                                                      & 59.0                                                          & 73.8                                                     & 81.2                        \\
\multicolumn{1}{c|}{PMG~\cite{du2020fine}}   &25M            & ResNet-50           & \cellcolor[HTML]{EFEFEF}89.6                        & \cellcolor[HTML]{EFEFEF}{\color[HTML]{000000} 93.4}  & 78.8                                                    & 91.0                                                      & 58.7                                                          & 72.2                                                     & 80.6                        \\
\multicolumn{1}{c|}{API-NET~\cite{zhuang2020learning}}    &36M       & ResNet-50           & \cellcolor[HTML]{EFEFEF}87.7                        & \cellcolor[HTML]{EFEFEF}93.0                         & 80.9                                                    & 93.6                                                      & 59.5                                                          & 74.3                                                     & 81.5                        \\
\multicolumn{1}{c|}{Cross-X~\cite{luo2019cross}}  &30M         & SENet-50            & \cellcolor[HTML]{EFEFEF}87.5                        & \cellcolor[HTML]{EFEFEF}92.7                         & 80.5                                                    & 93.1                                                      & 63.5                                   & {\color[HTML]{000000} 78.9}                              & 82.7                        \\
\multicolumn{1}{c|}{DCL~\cite{chen2019destruction}}     &28M          & ResNet-50           & \cellcolor[HTML]{EFEFEF}87.8                        & \cellcolor[HTML]{EFEFEF}93.0                         & 77.9                                                    & 90.2                                                      & 58.1                                                          & 70.9                                                     & 79.7                        \\
\multicolumn{1}{c|}{MGE~\cite{zhang2019learning}}    &28M            & ResNet-50           & \cellcolor[HTML]{EFEFEF}88.5                        & \cellcolor[HTML]{EFEFEF}90.8                         & 82.3                                                    & 93.0                                                      & 59.8                                                          & 74.5                                                     & 81.5                        \\
\multicolumn{1}{c|}{Mix+~\cite{li2020attribute}}   &25M           & ResNet-50           & \cellcolor[HTML]{EFEFEF}88.4                        & \cellcolor[HTML]{EFEFEF}92.0                         & 82.0                                                    & 92.8                                                      & 60.0                                                          & 74.8                                                     & 81.7                        \\
\multicolumn{1}{c|}{TransFG\cite{he2022transfg}}  &86M   & ViT-B\_16           & \cellcolor[HTML]{EFEFEF}91.7                        & \cellcolor[HTML]{EFEFEF}93.6  & 82.6                                                    & 94.8                                                      & 59.2                                                          & 78.5                                                     &  83.4 \\
\multicolumn{1}{c|}{RAMS-Trans~\cite{hu2021rams}}  &86M & ViT-B\_16           & \cellcolor[HTML]{EFEFEF}{ 91.3} & \cellcolor[HTML]{EFEFEF}92.7                         & 81.9                                                    & 93.2                                                      & 60.8                                                          & 78.2                                                     & 83.0                        \\
\multicolumn{1}{c|}{DualCross~\cite{zhu2022dual}} &88M   & ViT-B\_16           & \cellcolor[HTML]{EFEFEF}{\color[HTML]{32CB00} 92.0} & \cellcolor[HTML]{EFEFEF}93.3                         & 83.8                                                    & 94.3                                                      & 60.5                                                          & 78.8                                                     &  83.8 \\ \hline
\multicolumn{1}{c|}{DualStage~\cite{bajwa2019two}}  &33M       & UNet + ResNet-50    & 86.1                                                & 87.9                                                 & \cellcolor[HTML]{EFEFEF}80.3                            & 92.8                                                      & 59.7                                                          & -                                                        & -                        \\
\multicolumn{1}{c|}{DENet~\cite{fu2018disc}}     &30M         & ResNet-50           & -                                                   & -                                                    & \cellcolor[HTML]{EFEFEF}84.7                            & -                                                         & -                                                             & -                                                        & -                           \\
\multicolumn{1}{c|}{FundTrans~\cite{fan2022detecting}}  &86M   & ViT-B\_16           & 90.1                                                & 91.8                                                 & \cellcolor[HTML]{EFEFEF}85.3     & 93.5                                                      & 61.0                                                          & 78.1                                                     & 83.8                        \\
\multicolumn{1}{c|}{SatFormer~\cite{jiangsatformer}}    &92M      & ViT-B\_16           & 91.2                                                & 92.6                                                 & \cellcolor[HTML]{EFEFEF}85.0                            & 94.3                                                      & 62.3                                                          & 78.7                                                     & 84.0                           \\
\multicolumn{1}{c|}{Convid-ViT~\cite{zhang2021transformer}} &88M   & Swin-B              & -                                                   & -                                                    & -                                                       & \cellcolor[HTML]{EFEFEF}95.5                              & -                                                             & {\color[HTML]{3531FF} -}                                 & -                           \\
\multicolumn{1}{c|}{ConvidNet~\cite{ahmed2021convid}} &46M  & ConvidNet           & 83.2                                                & 81.0                                                 & 78.4                                                    & \cellcolor[HTML]{EFEFEF}93.3                              & 55.2                                                          & 67.5                                                     & 76.4                        \\
\multicolumn{1}{c|}{PSNet~\cite{tian2018psnet}}   &22M           & WRN                 & 86.7                                                & 87.9                                                 & 80.5                                                    & \cellcolor[HTML]{EFEFEF}94.0                              & 56.7                                                          & 69.0                                                     & 79.1                        \\
\multicolumn{1}{c|}{Convid-Trans~\cite{shome2021covid}}   &86M    & ViT-B\_16           & 90.1                                                & 91.2                                                 & 81.1                                                    & \cellcolor[HTML]{EFEFEF}95.0                              & 60.1                                                          & 77.6                                                     & 82.5                        \\
\multicolumn{1}{c|}{ResGANet~\cite{cheng2022resganet}}     &27M      & ResNet-50           & 86.7                                                & 90.0                                                 & \cellcolor[HTML]{EFEFEF}82.9                            & \cellcolor[HTML]{EFEFEF}94.0                              & 58.4                                                          & 76.5                                                     & 78.9                        \\
\multicolumn{1}{c|}{SynMIC~\cite{zhang2019medical}}    &30M         & ResNet-50           & 87.1                                                & 91.6                                                 & \cellcolor[HTML]{EFEFEF}83.6                            & \cellcolor[HTML]{EFEFEF}94.5                              & 59.8                                                          & 74.5                                                     & 81.9                        \\
\multicolumn{1}{c|}{SeATrans~\cite{wu2022seatrans}}  &96M    & UNet + ViT-B\_16    & 90.3                                                & 92.6                                                 & \cellcolor[HTML]{EFEFEF}{\color[HTML]{FE0000}87.6}                   & \cellcolor[HTML]{EFEFEF}{\color[HTML]{32CB00} 95.8}       & -                                                             & -                                                        & -                           \\ \hline
\multicolumn{1}{c|}{CLIP-Art~\cite{conde2021clip}}   &88M       & ViT-B\_32           & -                                                   & -                                                    & -                                                       & -                                                         & \cellcolor[HTML]{EFEFEF}60.8                                  & -                                                        & -                           \\
\multicolumn{1}{c|}{MLMO~\cite{gao2021multi}}     &27M     & ResNet-50           & 84.8                                                & 87.1                                                 & 77.8                                                    & 90.2                                                      & \cellcolor[HTML]{EFEFEF}60.3                                  & 75.2                                                     & 79.2                        \\
\multicolumn{1}{c|}{GCNBoost~\cite{el2021gcnboost}}    &21M      & GCN       & 85.2                                                & 88.9                                                 & 79.5                                                    & 92.6                                                      & \cellcolor[HTML]{EFEFEF}62.8                                  & 74.7                                                     & 80.6                        \\
\multicolumn{1}{c|}{Pavel~\cite{pavel}}   &155M    & SENet*2 + PNasNet-5 & -                                                   & -                                                    & -                                                       & -                                                         & \cellcolor[HTML]{EFEFEF}{\color[HTML]{FE0000}67.2}                         & -                                                        & -                           \\
\multicolumn{1}{c|}{DualPath~\cite{zhong2020fine}}     &29M     & ResNet-50           & 84.2                                                & 87.5                                                 & 77.8                                                    & 91.3                                                      & 60.1                                                          & \cellcolor[HTML]{EFEFEF}78.5                             & 79.9                        \\
\multicolumn{1}{c|}{Two-Stage~\cite{sandoval2019two}}    &48M     & ResNet-50           & 86.7                                                & 90.4                                                 & 83.9                                                    & 91.8                                                      & 62.7                                                          & \cellcolor[HTML]{EFEFEF}79.3                             & 82.5                        \\
\multicolumn{1}{c|}{DeepArt~\cite{mao2017deepart}}     &45M      & Vgg16               & -                                                   & -                                                    & -                                                       & -                                                         & 55.6                                                          & \cellcolor[HTML]{EFEFEF}75.2                             & -                           \\
\multicolumn{1}{c|}{RASA~\cite{lecoutre2017recognizing}}       &23M       & ResNet-34           & 81.8                                                & 85.2                                                 & 75.6                                                    & 87.0                                                      & 58.5                                                          & \cellcolor[HTML]{EFEFEF}76.3                             & 77.4                        \\
\multicolumn{1}{c|}{CrossLayer~\cite{chen2019recognizing}}   &42M     & Vgg16               & 81.1                                                & 84.7                                                 & 75.8                                                    & 86.8                                                      & 62.7                                                          & \cellcolor[HTML]{EFEFEF}77.0                             & 77.7                        \\ \hline
\multicolumn{1}{c|}{ResNet~\cite{he2016deep}}     &25M       & ResNet-50           & \cellcolor[HTML]{EFEFEF}84.5                        & \cellcolor[HTML]{EFEFEF}87.3                         & \cellcolor[HTML]{EFEFEF}77.3                            & \cellcolor[HTML]{EFEFEF}90.6                              & \cellcolor[HTML]{EFEFEF}59.2                                  & \cellcolor[HTML]{EFEFEF}74.1                             & 78.8                        \\
\multicolumn{1}{c|}{CvT~\cite{wu2021cvt}}    &32M      & CvT-21              & \cellcolor[HTML]{EFEFEF}90.6                        & \cellcolor[HTML]{EFEFEF}93.0                         & \cellcolor[HTML]{EFEFEF}82.1                            & \cellcolor[HTML]{EFEFEF}94.5                              & \cellcolor[HTML]{EFEFEF}61.5                                  & \cellcolor[HTML]{EFEFEF}79.7      & 83.6                        \\
\multicolumn{1}{c|}{DeiT~\cite{touvron2021training}}  &86M       & DeiT-B              & \cellcolor[HTML]{EFEFEF}91.1                        & \cellcolor[HTML]{EFEFEF}{\color[HTML]{32CB00} 93.8}  & \cellcolor[HTML]{EFEFEF}83.3                            & \cellcolor[HTML]{EFEFEF}95.7       & \cellcolor[HTML]{EFEFEF}60.7                                  & \cellcolor[HTML]{EFEFEF}{\color[HTML]{32CB00} 80.4}      & {\color[HTML]{32CB00} 84.2} \\
\multicolumn{1}{c|}{ViT~\cite{dosovitskiy2020image}}     &86M     & ViT-B\_16           & \cellcolor[HTML]{EFEFEF}90.3                        & \cellcolor[HTML]{EFEFEF}91.6                         & \cellcolor[HTML]{EFEFEF}81.8                            & \cellcolor[HTML]{EFEFEF}93.8                              & \cellcolor[HTML]{EFEFEF}60.4                                  & \cellcolor[HTML]{EFEFEF}78.8                             & 82.8                        \\ \hline
\multicolumn{1}{c|}{ExpNet-S}  &16M   & ResNet18 + CA       & \cellcolor[HTML]{EFEFEF}89.4                        & \cellcolor[HTML]{EFEFEF}92.8                         & \cellcolor[HTML]{EFEFEF}{\color[HTML]{000000} 84.8}     & \cellcolor[HTML]{EFEFEF}95.1                              & \cellcolor[HTML]{EFEFEF}{\color[HTML]{000000} 61.8}           & \cellcolor[HTML]{EFEFEF}78.5                             & 83.7                        \\
\multicolumn{1}{c|}{ExpNet-M}  &47M   & ResNet50 +CA        & \cellcolor[HTML]{EFEFEF}{\color[HTML]{3531FF} 92.6} & \cellcolor[HTML]{EFEFEF}{\color[HTML]{3166FF} 94.2}  & \cellcolor[HTML]{EFEFEF}{\color[HTML]{32CB00} 86.8}     & \cellcolor[HTML]{EFEFEF}{\color[HTML]{3166FF} 96.7}       & \cellcolor[HTML]{EFEFEF}{\color[HTML]{32CB00} 64.2}           & \cellcolor[HTML]{EFEFEF}{\color[HTML]{3531FF} 81.9}      & {\color[HTML]{3166FF}86.1}                        \\
\multicolumn{1}{c|}{ExpNet-L} &64M    & ResNet50 +CA        & \cellcolor[HTML]{EFEFEF}{\color[HTML]{FE0000} 92.8}              & \cellcolor[HTML]{EFEFEF}{\color[HTML]{FE0000}94.6}                & \cellcolor[HTML]{EFEFEF}{\color[HTML]{3166FF} 87.2}     & \cellcolor[HTML]{EFEFEF}{\color[HTML]{FE0000}96.9}                     & \cellcolor[HTML]{EFEFEF}{\color[HTML]{3166FF} 65.7}           & \cellcolor[HTML]{EFEFEF}{\color[HTML]{FE0000}83.1}                    & {\color[HTML]{FE0000}86.7}               \\ \hline \bottomrule
\end{tabular}}
\caption{The comparison of ExpNet with SOTA classification methods in different fields. The gray background denotes the method is proposed for that specific task/tasks. The metric Acc denotes accuracy, and AUC denotes area under the ROC curve. The $1^{st}$, $2^{nd}$ and $3^{rd}$ methods are denoted as {\color[HTML]{FE0000}red}, {\color[HTML]{3166FF}blue} and {\color[HTML]{32CB00}green} respectively. The number of the parameters is counted in million (M).}\label{tab:mainres}
\vspace{-15pt}
\end{table*}
\vspace{-3pt}
\subsection{Dataset}
We conduct the experiments on three representative expert-level classification tasks: FGVC, medical image classification (diagnosis), and art attributes classification. For FGVC, two commonly used benchmarks, CUB-200-2011 (CUB)~\cite{wah2011caltech} and FGVC-Aircraft (Air)~\cite{maji2013fine} are involved in demonstrating the performance of our method. In medical image classification, we use Convidx dataset~\cite{wang2020covid}, which is a large-scale and open access benchmark dataset for predicting COVID-19 positive cases from chest X-Ray images, and REFUGE2 dataset~\cite{fang2022refuge2}, which is a publicly released challenge dataset for screening glaucoma from fundus images. WikiArt~\cite{saleh2015large} and iMet~\cite{zhang2019imet} are used as two art-related datasets. WikiArt consists of paintings and iMet mainly consists of artworks in The Metropolitan Museum like sculptures or porcelains. 
\vspace{-3pt}
\subsection{Setting}
We experiment with the large, medium, and small variants of our model, \textit{ExpNet-L}, \textit{ExpNet-M}, and \textit{ExpNet-S}, respectively. In \textit{ExpNet-S}, we use ResNet18 as the backbone and use MLP + addition for the embedding fusion. In \textit{ExpNet-M} and \textit{ExpNet-S}, we use ResNet50 as the backbone and CA for the embedding fusion. \textit{ExpNet-M} adopts CA with $12$ heads and $768$ hidden size. \textit{ExpNet-L} adopts CA with $16$ heads and $1024$ hidden size. On all variant models, the images are uniformly resized to the dimension of 448$\times$448 pixels. We train the models for 100 epochs using AdamW\cite{kingma2014adam} with batch size of 16. Detailed configurations and training settings are provided in the supplementary material.
\begin{figure}[!t]
\centering
\includegraphics[width=\linewidth]{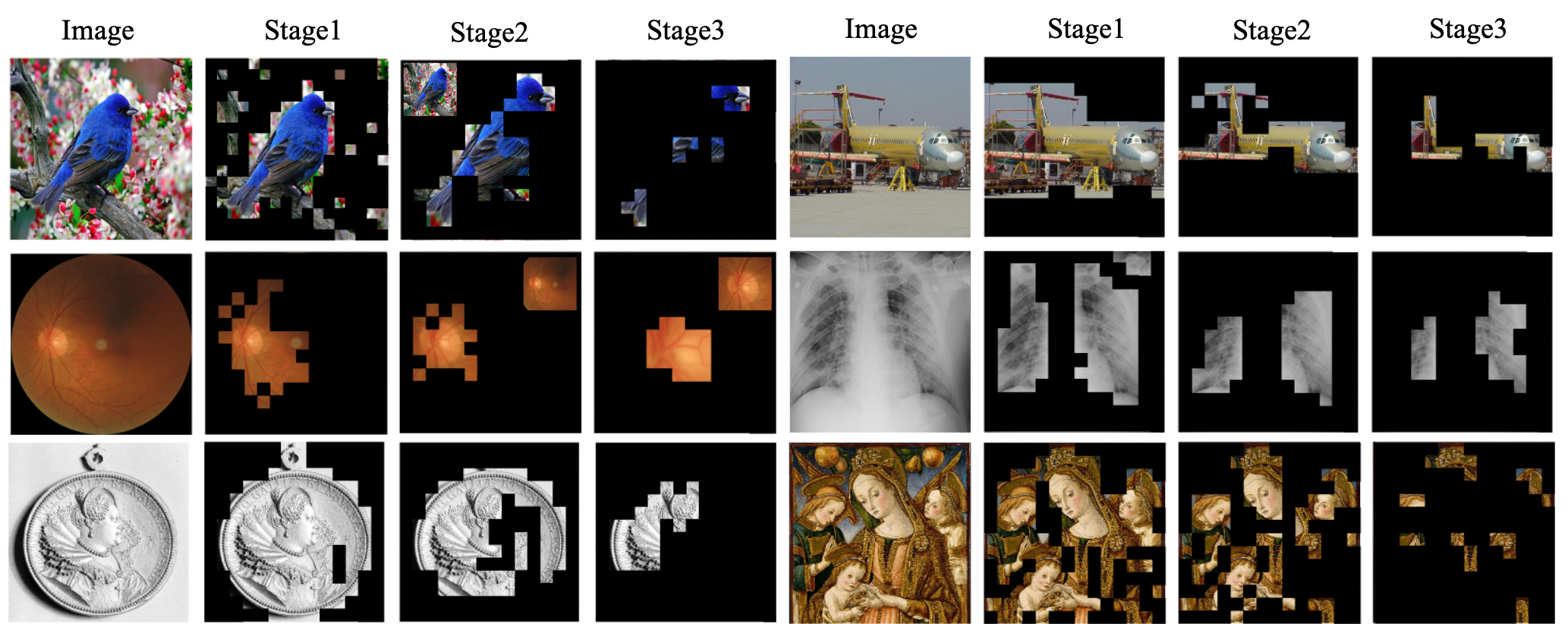}
\caption{The visualized results of \textit{ExpNet-M} saliency map mapping back to the raw image in each stage (from shallow to deep, denoted as stage 1, 2, and 3). The focal parts are kept and the context parts are masked. For the small focal parts, we zoom in the region on the up-left/right corners for clarity. From top to down are FGVC, medical image, and artwork classification respectively.}
\label{fig:mask_vis}
\vspace{-15pt}
\end{figure}

\vspace{-3pt}
\subsection{Main Results}
We compare ExpNet with SOTA classification methods proposed in four different fields: FGVC, medical image classification, artwork attribution recognition, and general vision classification. The main results are shown in Table~\ref{tab:mainres}. 
For the better analysis of ExpNet, we also show the visualized results of the intermediate saliency maps in Fig. \ref{fig:mask_vis}. The maps are mapping back to the raw image space for the analysis.
\vspace{-3pt}
\subsubsection{ExpNet \textit{vs.} FGVC methods}
In FGVC, a typical strategy is to identify the intra-class and inter-class variances, like BCN~\cite{dubey2018pairwise}, API-NET~\cite{zhuang2020learning}, PMG~\cite{du2020fine} and DCL~\cite{chen2019destruction}. 
But without the attention to the discriminative parts, they do not perform well on FGVC and medical images, in which the regional lesions/details are vital for the correct classification. 
MGE~\cite{zhang2019learning}, Mix+~\cite{li2020attribute}, TransFG~\cite{he2022transfg} and DualCross~\cite{zhu2022dual} select and reinforce the discriminative parts for FGVC; however, they either ignore the context information~\cite{he2022transfg,zhang2019learning} or equally process the local-global features~\cite{zhu2022dual,li2020attribute}, which limited their performance.



As CUB and Air cases shown in Fig.~\ref{fig:mask_vis}, the proposed ExpNet also focuses on the discriminative details, like part-based FGVC methods. The difference is that ExpNet also individually models the Context Impression for the classification. Thanks to individual part-context modeling, \textit{ExpNet-L} outperforms SOTA DualCross by 0.8\% and 1.3\% on CUB and Air, respectively.

\vspace{-3pt}
\subsubsection{ExpNet \textit{vs.} Medical Imaging Methods}
Medical image classification tasks need to focus on both the local lesions and the global structural relationship of the organs/tissues. For example, in the glaucoma screening, the network needs to first locate the optic-disc region\cite{bajwa2019two} on the fundus image and then model the relationship between optic disc and optic cup\cite{fu2018disc} for the diagnosis.
On CONVID-19 diagnosis, methods are also designed to find and focus on the airspace opacities from the chest X-Ray images\cite{ahmed2021convid,tian2018psnet,zhang2021transformer}. However, they often can not model the global structure, thus performing worse for the glaucoma classification. 
The general medical image diagnosis methods~\cite{cheng2022resganet,wu2022seatrans,zhang2019learning} are often designed with both regional attention and feature interaction. 
For example, ResGANet~\cite{cheng2022resganet} uses spatial attention and channel attention respectively to reinforce the discriminative features. SeATrans~\cite{wu2022seatrans} combines the localization UNet and classification network at the feature-level through ViT.

Unlike these methods, ExpNet abstracts the discriminative region and models the global structure relationship in a unified network. As the glaucoma case shown in Fig.~\ref{fig:mask_vis}, ExpNet focuses on the optic disc in Stage2 and the optic cup in Stage3, which we can infer the fusion of the focal embedding and the last Complex Impression can effectively model the optic-disc/cup relationship (verified in Section \ref{sec:analysis}).
The results show \textit{ExpNet-L} ranks the first on CONVID-19 and the second on REFUGE2. Our performance on glaucoma diagnosis is even competitive with the segmentation-assisted method (SeATrans), without using any prior segmentation.
\vspace{-3pt}
\subsubsection{ExpNet \textit{vs.} Artwork recognition methods}
On the art-related dataset, modeling the interaction of the hierarchical features is an effective strategy\cite{chen2021crossvit,sandoval2019two}. That is because predicting the attributions of the artwork, e.g., the artist or the culture, needs a comprehensive interaction of various features, like the style or stroke of the painting, the pattern of the ornaments on pottery, etc. Motivated by this observation, CrossLayer\cite{chen2021crossvit} applies the interactions on the high-level layers. Two-Stage\cite{sandoval2019two} first splits the image by overlapped windows and separately extracts their features, and uses another CNN to learn the interaction of the high-level features, achieving the highest performance among the CNN-based models.

In ExpNet, NeFiRF can effectively recognize orthogonal features for each stage, and thus facilitate the final feature interaction. As the painting case shown in Fig.~\ref{fig:mask_vis}, ExpNet individually models the background, the figure faces/gestures, and ornament patterns in the three stages. After abstracting the different features at each stage, the final interaction will become more effective. As shown in Table \ref{tab:mainres}, \textit{ExpNet-L} ranks the first on WikiArt and the second on iMet. The only method ranking higher (Pavel) uses double the parameters.

\vspace{-3pt}
\subsubsection{ExpNet \textit{vs.} General vision classification}
We also compare our method with various general vision classification architectures. We take ResNet50, two modified ViT architectures~\cite{wu2021cvt,touvron2021training} and vanilla ViT~\cite{dosovitskiy2020image} for the comparison. We can see these methods are often more general, but commonly perform worse than SOTA on the specific tasks. 
The proposed ExpNet performs well on all three tasks, and shows the best generalization ability. Comparing the mean score with general vision classification methods, \textit{ExpNet-M} outperforms SOTA DeiT by 2\% with half of the parameters.
\vspace{-3pt}
\subsubsection{Cross-dataset Generalization}
Considering the cross-dataset generalization performance of different methods, we can see that the part-based strategies often work well on both fine-grained and medical image classification. For example, MGE, TransFG, DualCross, SatFormer, SeATrans often work well on both FGVC and medical tasks. In these tasks, the local parts, like bird beads or optic-cup are very important for the classification.
The methods modeling the interactions of hierarchical features are likely to perform better on art-related datasets. For example, Cross-X shows strong generalization on art-related dataset.  
We can also see that the stronger backbone architecture will be more robust. Considering the mean performance, the top methods in each field are commonly based on the ViT backbone. The proposed ExpNet combines part-context modeling, hierarchical feature interaction, and appropriate network architectures in a unified framework, which achieves top-3 performance in all the tasks and achieves the best mean score with a good accuracy-complexity trade-off.


\begin{figure}[!t]
\centering
\includegraphics[width=\linewidth]{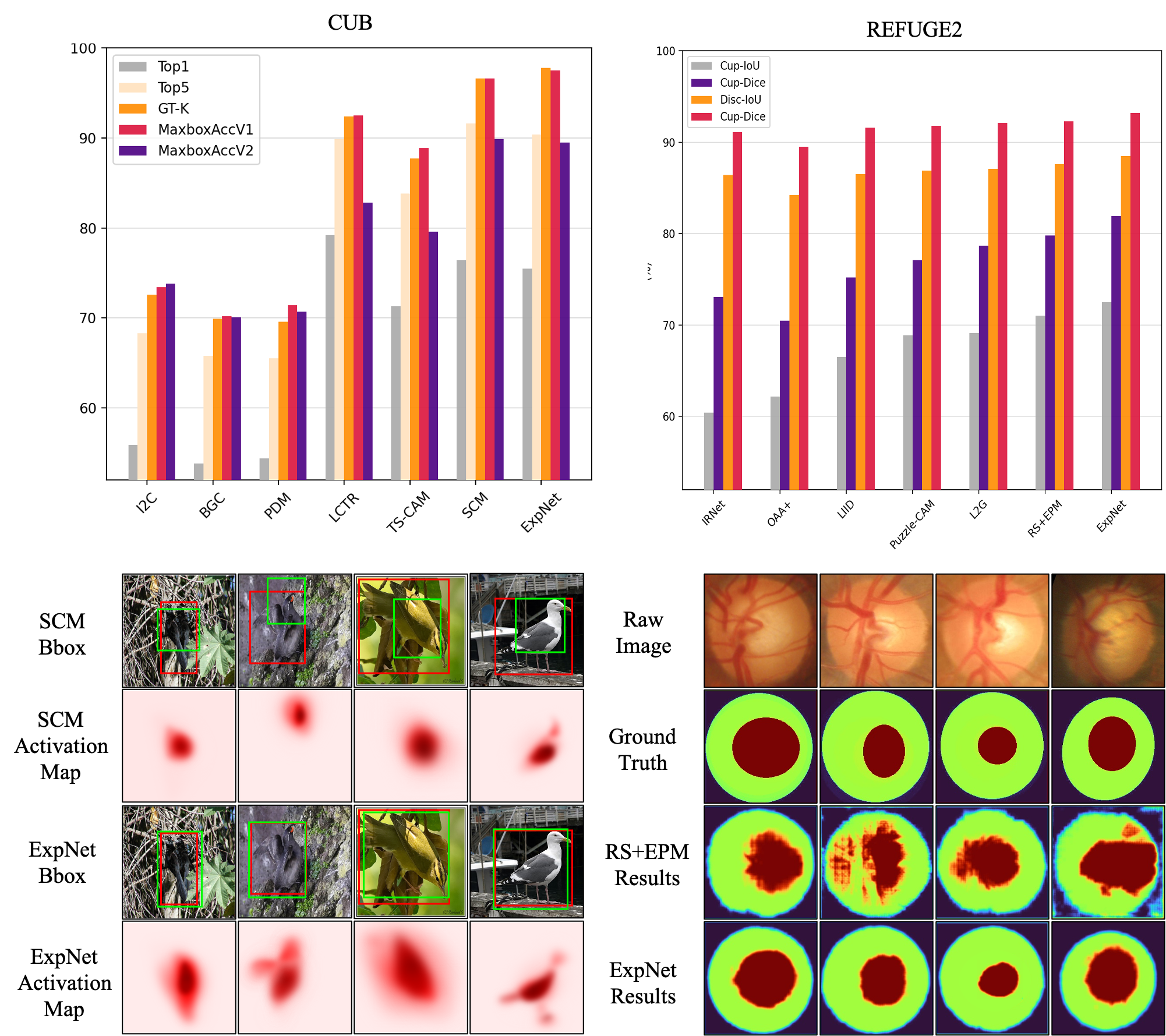}
\caption{Quantitative and qualitative comparison of weakly supervised detection and segmentation on CUB and REFUGE2, respectively. On CUB, we show activation score map and bounding box (Bbox) of SCM for comparison. The ground-truth and predicted Bboxes are shown as {\color[HTML]{FE0000}red} and {\color[HTML]{32CB00}green} respectively. On REFUGE2, we show optic disc ({\color[HTML]{32CB00}green}) and cup ({\color[HTML]{800020}brown}) segmentation of RS+EPM for the comparison. 
}
\label{fig:increment}
\vspace{-15pt}
\end{figure}

\vspace{-3pt}
\subsection{Weakly-supervised Localization/Segmentation}\label{sec:exp_prior}
We can observe from Fig.~\ref{fig:mask_vis} that ExpNet has a close relationship with the object shape and location. In the section, we further explore this relationship through quantitative and qualitative experiments, which show ExpNet is also a strong model for weakly-supervised localization and segmentation. 

To demonstrate the effect of weakly-supervised learning, we use the intermediate saliency map $m$ to produce object segmentation or localization results over \textit{ExpNet-M}. We produce the bird localization prediction from the first-stage map $m^{1}$ on CUB dataset, and produce the optic disc and cup segmentation prediction from $m^{2}$ and $m^{3}$ respectively on REFUGE2 dataset (details are in the supplementary material). On CUB dataset, we compare against the mainstream weakly-supervised localization methods, including l2C~\cite{zhang2020inter}, BGC~\cite{kim2022bridging}, PDM~\cite{meng2022diverse}, LCTR~\cite{chen2022lctr}, TS-CAM~\cite{gao2021ts}, and SCM~\cite{bai2022weakly}. We evaluate the performance by the commonly used GT-Known (GT-K), Top1/Top5 Localization Accuracy, and more strict ones like MaxboxAccV1 and MaxboxAccV2~\cite{choe2020evaluating}. On REGUGE2 dataset, we compare against various weakly-supervised segmentation methods, including IRNet~\cite{ahn2019weakly}, OAA++~\cite{jiang2019integral}, LIID~\cite{liu2020leveraging}, Puzzle-CAM~\cite{jo2021puzzle}, L2G~\cite{jiang2022l2g}, and RS+EPM~\cite{jo2022recurseed} through IoU and Dice Score. We show the quantitative results and visual comparison in Fig.~\ref{fig:increment}. On the CUB dataset, ExpNet outperforms SOTA SCM with 1.2\% over GT-K and 1.1\% over MaxboxAccV1. The visualized results show that ExpNet predicts more sophisticated activation maps, and thus produces more accurate bounding boxes. On REFUGE2 dataset, ExpNet outperforms SOTA RS-EPM with 2.5\% over IoU and 2.1\% over Dice. Compared with RS+EPM over the visualized results, ExpNet predicts neater and more reasonable segmentation masks, especially on the ambiguous optic-cup.
\vspace{-3pt}
\subsection{Ablation Study and Alternatives}
In order to verify the effectiveness of the proposed NeFiRF, we compare it with alternative feature grouping/activation techniques, including Spatial Attention~\cite{woo2018cbam}, Group Trans~\cite{liu2022dynamic} and Global Filter~\cite{rao2021global}. The experimental results over \textit{ExpNet-M} are shown in Table \ref{tab:ab_freq}. We can see that the frequency-based method Global Filter, basically performs better than space-based methods Spatial Attention and Group Trans. 
The proposed NeFiRF which introduces spatial constraint in the frequency space outperforms Global Filter by 0.6\%, 0.9\%, and 1.8\% on three datasets. Compared with the baseline NeRF, NeFiRF significantly improves 1.2\%, 2.3\% and 2.0\% on three datasets, demonstrating the effectiveness of the proposed conditional sine activation and band-pass filter.

To further verify the effectiveness of each proposed component, we perform detailed ablation studies over both Gaze-Shift and NeFiRF on \textit{ExpNet-M}, as listed in Table \ref{tab:ablation}. In Table \ref{tab:ablation}, we sequentially add the proposed modules on top of the ResNet50 baseline, and the model performance is gradually improved. First, we adopt Focal Part separation over the baseline with Spatial Attention (SA) and CNN-based spatial organized (Section \ref{sec:so}) context embedding. It increases the performance on the tasks in which certain parts are discriminative (CUB and COVID-19). By applying part-context correlated CA Context Impression, the performance is significantly increased, indicating that global attention is more effective for the context information modeling. Then we use NeFiRF with only conditional sine activation to replace SA, the performance again significantly improved. According NeRF results in Table \ref{tab:ab_freq}, we can infer conditional sine activation individually improves 0.9\%, 2.2\% and 2.2\% on three datasets. Further applying band-pass filters over NeFiRF increases the model performance over all three datasets, which demonstrates the general effectiveness of this simple regularization strategy.

\begin{table}[!t]
\centering
\caption{Comparison of NeFiRF and alternative feature grouping or activation strategies.}
\resizebox{0.45\textwidth}{!}{
\begin{tabular}{c|ccclc}
\toprule
\hline
          & \begin{tabular}[c]{@{}c@{}}Spatial \\ Attention\end{tabular} & \begin{tabular}[c]{@{}c@{}}Group\\ Trans\end{tabular} & \begin{tabular}[c]{@{}c@{}}Global \\ Filter\end{tabular} & NeRF                       & \begin{tabular}[c]{@{}c@{}}NeFiRF\end{tabular} \\ \hline
CUB        & 91.9                                                      & {\color[HTML]{000000} 92.0}                             & 91.7                                                    & {\color[HTML]{000000} 91.4} &  92.6                                                       \\
Convid-19  & 94.6                                                      & 95.2                                                    & 95.6                                                   & 94.4                        &  96.7                                                       \\
WikiArt    & 79.5                                                      & 79.9                                                    & 80.8                                                   & 79.1                        &  81.9                                                       \\ \hline \bottomrule
\end{tabular}}\label{tab:ab_freq}
\vspace{-5pt}
\end{table}

\begin{table}[!t]
\centering
\caption{Ablation study over Gaze-Shift and NeFiRF.}
\resizebox{0.47\textwidth}{!}{
\begin{tabular}{c|cc|cc|ccc}
\toprule
\hline
Baseline & \multicolumn{2}{c|}{Gaze-Shift}                                                                                & \multicolumn{2}{c|}{NeFiRF}                                            & CUB  & COVID-19 & WikiArt \\ \hline
         & \begin{tabular}[c]{@{}c@{}}Focal\\ Part\end{tabular} & \begin{tabular}[c]{@{}c@{}}Context\\ Impression\end{tabular} & \begin{tabular}[c]{@{}c@{}}Cond Sine\\ Activation\end{tabular} & \begin{tabular}[c]{@{}c@{}}Band-Pass\\ Filter\end{tabular} &      &       &      \\ \hline
\checkmark     &                                                           &                                                      &                                                               &           & 84.5 & 90.6  & 74.1 \\
\checkmark     & \checkmark                                                      &   CNN                                                   &     \multicolumn{2}{c|}{SA}           & 87.2 & 92.9  & 74.8 \\
\checkmark     & \checkmark                                                      & \checkmark                                                 &   \multicolumn{2}{c|}{SA}                                                                       & 91.9 & 94.6  & 79.5 \\
\checkmark     & \checkmark                                                      & \checkmark                                                 & \checkmark                                                          &           & 92.3 & 96.2  & 81.3 \\
\checkmark     & \checkmark                                                      & \checkmark                                                 & \checkmark                                                          & \checkmark      & 92.6 & 96.7  & 81.9 \\ \hline \bottomrule
\end{tabular}}\label{tab:ablation}
\vspace{-5pt}
\end{table}

\begin{figure}[h]
    \centering
    \includegraphics[width=0.95\linewidth]{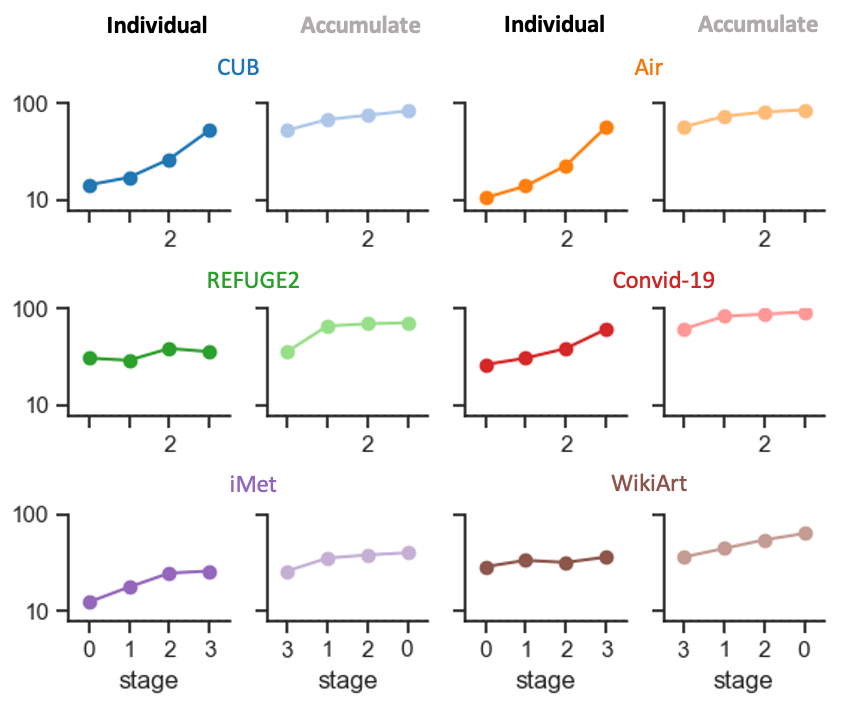}
    \vspace{-5pt}
    \caption{The classification performance of the each-stage intermediate result. Stages 0, 1, 2, 3 correspond to the first, second, third Context Impression and final focus embedding, respectively. Individual indicates the performance of each stage alone. Accumulation represents the performance after sequential fusion.}
    \label{fig:ana_line}
    \vspace{-15pt}
\end{figure}

\vspace{-3pt}
\subsection{Analysis and Discussion}\label{sec:analysis}
For an in-depth understanding of the function and effect of each stage in ExpNet, 
we quantitatively test the performance of Context Impression produced in each stage. Specifically, we connect and train a fully-connected layer on each embedding of a frozen \textit{ExpNet-M} to predict the classification, and use the same way to test the performance of each sequential fusion of the embedding.
The results are denoted as Individual and Accumulate respectively in Fig.~\ref{fig:ana_line}. 
We can see that for the tasks in which the local parts are discriminative, like CUB, Air, and Convid-19, the deeper embeddings often contribute more to the right decision. 
On REFUGE2, we can see from the Accumulate results that the fusion of the last two embeddings are the most important. That is large because the last two embeddings represent optic disc and cup parts respectively (as shown in Fig.~\ref{fig:mask_vis}), and their fusion will learn the key parameters, i.e., vCDR for the final classification. On WikiArt, the fusion linearly improves the performance. We can infer that various features, like image style, painting stroke, and object patterns, are all critical for discrimination. Based on the above analysis, we can see that the discriminant factors vary greatly in different expert level classification tasks. The previous methods, like FGVC, will be hard to cover all these factors. But we design ExpNet to fit different discrimination manners in a unified architecture, which achieves the best performance in expert-level classification.
\vspace{-15pt}
\section{Conclusion}
In this work, we focused on a particular vision classification task requiring professional categories, i.e., expert-level classification. As the task is hard to be addressed with the existing solutions, we proposed to hierarchically decouple the part and context features in our ExpNet and individually modeled the two parts through different architectures. It enables the network to focus on the discriminative parts and perceive the context content in a unified framework, as well as to integrate hierarchical features for comprehensive predictions.
Extensive experiments demonstrated the overall superior performance of our ExpNet on a wide range of expert-level classification tasks.

{\small
\bibliographystyle{ieee_fullname}
\bibliography{egbib}
}

\clearpage

\end{document}